\documentclass[conference]{IEEEtran}
\usepackage{times}

\usepackage[numbers]{natbib}
\usepackage{multicol}
\usepackage[bookmarks=true]{hyperref}
\usepackage[dvipsnames]{xcolor}
\usepackage{graphicx}
\usepackage{microtype}

\hypersetup{
linkcolor=BrickRed
,citecolor=Green
,filecolor=Mulberry
,urlcolor=NavyBlue
,menucolor=BrickRed
,runcolor=Mulberry
,linkbordercolor=BrickRed
,citebordercolor=Green
,filebordercolor=Mulberry
,urlbordercolor=NavyBlue
,menubordercolor=BrickRed
,runbordercolor=Mulberry
,colorlinks=True
}

\begin{document}

\title{Bio-inspired Robot Perception\\Coupled With Robot-modeled Human Perception}

\author{Tobias Fischer%
}

\maketitle

\IEEEpeerreviewmaketitle

\begin{figure}
    \centering
    \includegraphics[width=0.99\linewidth]{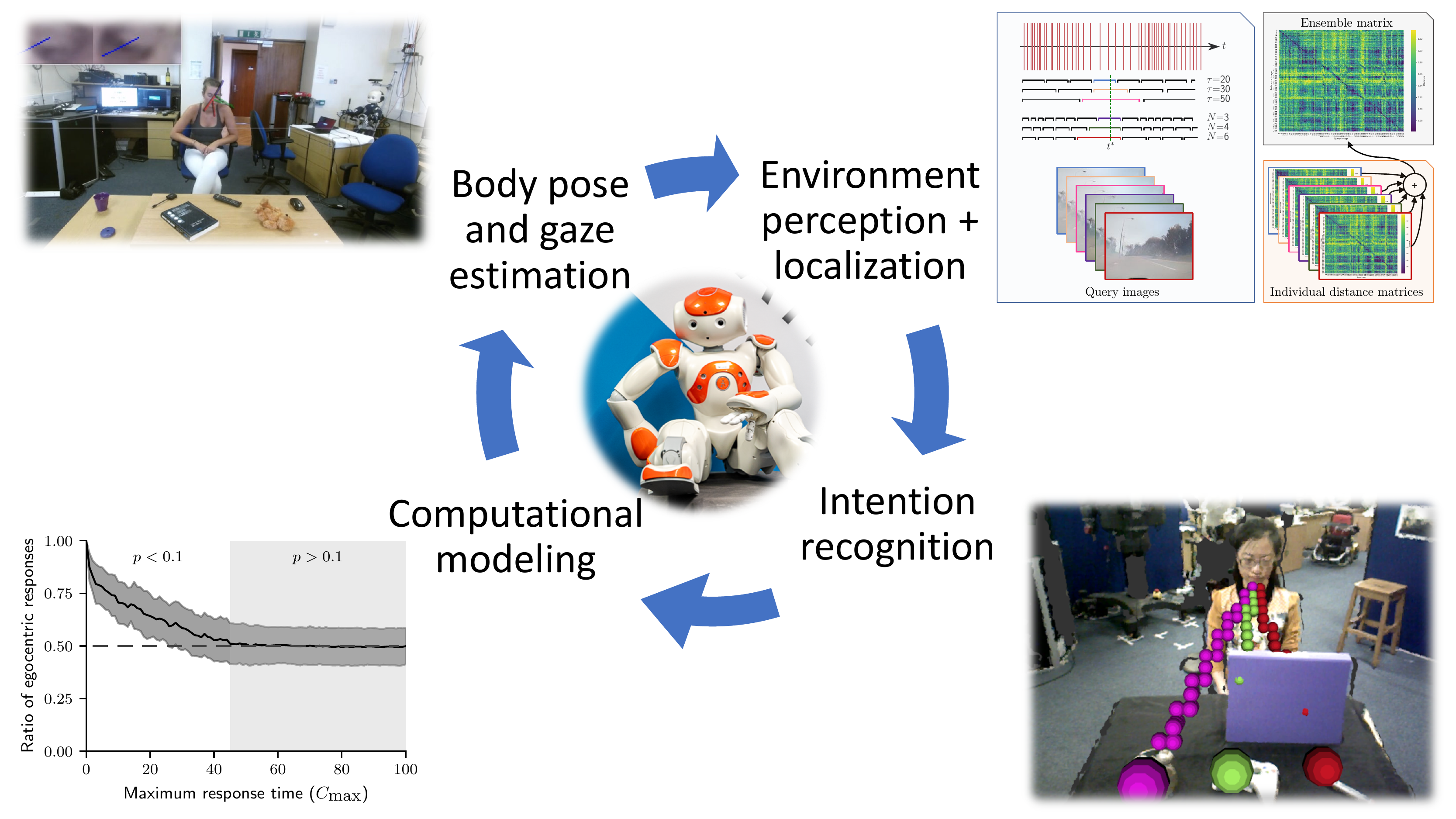}
    \caption{My research is centered around human perception (including body pose and gaze estimation), environment perception (including localization and navigation), higher-level human understanding (including intention recognition) and computational modeling to gain insights into the underlying mechanisms of visual perception in humans.}
    \vspace{-\baselineskip}
    \label{fig:main}
\end{figure}

\textbf{Research statement -- part 1:} My overarching research goal is to provide robots with perceptional abilities that allow interactions with humans in a human-like manner. To develop these perceptional abilities, I believe that it is useful to study the principles of the human visual system. I use these principles to develop new computer vision algorithms and validate their effectiveness in intelligent robotic systems. I am enthusiastic about this approach as it offers the dual benefit of uncovering principles inherent in the human visual system, as well as applying these principles to its artificial counterpart. Fig.~\ref{fig:main} contains a depiction of my research.

\textbf{Perspective-taking:} In our everyday lives, we often interact with other people. Although each interaction is different and hard to predict in advance, they are usually fluid and efficient. This is because humans take many aspects into account when interacting with each other: the relationship between the interlocutors, their familiarity with the topic, and the time and location of the interaction, among many others. More specifically, humans are remarkably good at rapidly forming models of others and adapting their actions accordingly. To form these models, humans exploit the ability to take on someone else's point of view -- they \emph{take their perspective}~\cite{Fischer2018PhD}. 

In \cite{Fischer2016}, we introduced an artificial visual system that equips an iCub humanoid robot with the ability to perform perspective-taking in unknown environments using a depth camera mounted above the robot, i.e.\ without using a motion capture system or fiducial markers. Grounded in psychological studies~\cite{Michelon2006,Flavell1981}, perspective-taking is separated into two processes: level 1 perspective-taking comprises the ability to identify objects which are occluded from one perspective but not the other; this was implemented using line-of-sight tracing. Level 2 perspective-taking refers to understanding \emph{how} the object is perceived from the other perspective (rather than just understanding \emph{what} is visible from that perspective; see \cite{Michelon2006}); this was implemented using a mental rotation process.

While~\cite{Fischer2016} implements perspective-taking in a robotic system, it does not provide any insights into the underlying mechanisms used by humans. In \cite{FischerTCDS2020}, we investigated possible implementations of perspective-taking in the human visual system using a computational model applied to a simulated robot. The model proposes that a mental rotation of the self, also termed \emph{embodied transformation}, accounts for this ability. The computational model reproduces the reaction times of human subjects in several experiments and explains gender differences that were observed in human subjects.

In future works, I would like to explore the relationship between perspective-taking and active vision. Active vision refers to the manipulation of the robot's viewpoint to extract more information from their environment, given a specific task~\cite{Bajcsy2018}. I argue that visual perspective-taking is a subset of the active vision task, where the particular task is to gain an understanding of the other person's viewpoint, while minimizing the movements required to gain this understanding.

\textbf{Blink and gaze estimation: }Perspective-taking requires the robot to follow the gaze of the person that it is interacting with. While the computer vision community has made significant progress in the area of gaze estimation, the application is typically limited to settings where the head pose is constrained, and the person is close to the camera; examples include phone~\cite{Krafka2016} and tablet~\cite{Huang2017} use, as well as screen-based settings~\cite{Zhang2019}. However, scenarios that are typically encountered in human-robot interactions have seen little attention~\cite{lanillos2017bayesian,Schillingmann2015,palinko2015eye}.

To overcome this challenge, we proposed RT-GENE~\cite{FischerECCV2018}, a new dataset for gaze estimation in unconstrained settings. Opposed to other datasets that require the subject to gaze at a particular point and indicate once they have done so (e.g.~by pressing the space bar), \emph{RT-GENE's data labeling is fully automated}. The subject wears eyetracking glasses that are equipped using motion capture markers, which enables to simultaneously collect head pose and eye gaze data -- even in settings where the subject is far away (indeed even if they look away) from the camera.

However, the eyetracking glasses introduce an object at training time that is not present at inference time. To circumvent this issue, RT-GENE makes use of Generative Adversarial Networks (GANs) to inpaint the area covered by the eyetracking glasses. In particular, we collect a second dataset where the eyetracking glasses are not worn (but no gaze information is available) to train the GAN, and find the area to be inpainted by projecting down the known 3D model onto the image. RT-GENE also introduces a novel deep network that outperforms the state-of-the-art using an ensemble scheme. Interestingly, a deep model trained on RT-GENE performs exceptionally well on other datasets without any further adaptation. RT-GENE was tested in real-life settings: for example, the estimated gaze was used as the input signal on a powered wheelchair, and \emph{enabled a user to control the wheelchair purely based on their gaze}.

The RT-GENE dataset was also re-purposed for the blink estimation task to provide a challenging new dataset for blink estimation~\cite{Cortacero2019}. Using a similar ensemble scheme as for RT-GENE, along with an oversampling technique (there are significantly more images with open eyes than with closed eyes), an ensemble of deep networks was trained. The ensemble was then evaluated on various blink datasets where we outperformed prior methods by a large margin.

In future works, I would like to extend RT-GENE to work in even more challenging scenarios, where the eyes might not be visible at all. Incorporating the saliency of the scene, past gaze behavior, and other modalities like audio are interesting directions to tackle these scenarios, which will also require estimating the uncertainty of the gaze predictions, for example via the agreement of the ensemble members~\cite{lakshminarayanan2016simple}.

\textbf{Research statement -- part 2:} Part 1 of this paper introduced a variety of perceptional algorithms for robots -- blink estimation, gaze estimation and perspective-taking -- and how they can be used to uncover potential mechanisms of human perception. The following part 2 is concerned with localization and navigation capabilities of mobile robotic systems. As in part 1, we take inspiration from the principles discovered in the animal kingdom to develop new computer vision algorithms that are validated in mobile robotic systems~\cite{DallOsto2020,Fischer2020RAL,FischerGargIJCAI2021,Hausler2021}. Indeed, even ``simple'' animals like ants and rats can rapidly explore, localize and navigate in both familiar and new environments, despite their low visual acuity.

One key component of localization is visual place recognition, which refers to the problem of retrieving the reference image that is most similar to the query image, given a reference dataset of places that have been previously visited and a query image of the current location~\cite{FischerGargIJCAI2021}. Visual place recognition is a challenging problem, as query and reference images could be taken at different times of the day, in varying seasonal and weather conditions, and even years apart and with significant viewpoint changes.

\textbf{Demystifying Visual Place Recognition (VPR): }VPR is a rapidly growing topic, with over 2300 papers listed in Google Scholar matching the exact term. As VPR is researched in multiple communities with a wide range of downstream tasks, VPR research has become increasingly dissociated. For example, the evaluation metrics and datasets vary significantly across publications, which makes a comparison of methods challenging. In a recent survey~\cite{FischerGargIJCAI2021}, we highlight similarities and differences of VPR with a handful of related areas (including image retrieval, landmark recognition and visual overlap detection), and introduce three key drivers of VPR research: the platform, the operating environment, and the downstream task. Having gained this knowledge, this then enabled us to introduce and highlight several open research problems: how places should be represented (e.g.~can geometric information be incorporated into global image descriptors, how can descriptors be compressed efficiently, and can we synthesize novel views) and how places should be matched (e.g.~using hierarchical systems or learning to match). As with all of my research, I put an emphasis on what we can learn from biology -- in this example, the definition of a `place' is based on the spatial viewpoint cells that were found in the primate brain.

\textbf{Event-based VPR: }In~\cite{Fischer2020RAL}, we argue that it is beneficial to use event cameras to sense the environment, as opposed to conventional frame-based cameras. Event cameras are bio-inspired sensors that output a near-continuous stream of events; each event is denoted by the time $t$ and pixel location $(x, y)$ where the event was observed, as well as the event's polarity $p$ that indicates whether the intensity at the pixel decreased ($p=-1$) or increased ($p=+1$)~\cite{Gallego2019}. Therefore, event cameras only observe \emph{changes} within the environment, and an ideal event camera does not output any events in the case of a static environment. Event cameras have a range of advantages: they are so fast that no motion blur occurs, they have a very high dynamic range, and they are energy-efficient. 

We first reconstructed frames from the event stream using~\cite{Rebecq19pami}, and then applied conventional VPR image retrieval techniques~\cite{Arandjelovic2018}. The main novelty is in the use of multiple image sets, whereby each image set is obtained by reconstructing the images based on a different temporal scale. The overall best matching image is then found using an ensemble scheme, which is a recurring theme of my work. We are expanding this work to avoid the computationally expensive reconstruction, and instead directly use the event stream. We are also planning to make use of the recent advances in neuromorphic engineering~\cite{davies2021advancing} and deploy these algorithms on spiking neural networks implemented on neuromorphic hardware.

\textbf{Locally-global features for VPR:} In another line of work, we introduce Patch-NetVLAD~\cite{Hausler2021}, a novel formulation for combining the advantages of both local and global descriptor methods. NetVLAD~\cite{Arandjelovic2018} is a popular deep-learned global descriptor method that extracts a feature representation given an image that can be compared to another image's feature representation by means of standard distance metrics like the Euclidean distance. In Patch-NetVLAD~\cite{Hausler2021}, rather than aggregating across the whole feature space, we aggregate \emph{locally-global descriptors} from a set of patches in the feature space. This is implemented in a computationally efficient integral feature space, which further enables a multi-scale approach to fuse multiple patch sizes. Patch-NetVLAD outperforms the state-of-the-art by large margins on seven datasets. In future works, we would like to explore whether Patch-NetVLAD's multi-scale approach has an analogous mechanism in the animal kingdom, noting that the brain processes visual information over multiple receptive fields.

\addtolength{\textheight}{-1cm}

\textbf{Acknowledgments:} I would like to thank all my collaborators over the last years who have made this research possible, in particular my supervisors \href{https://www.demiris.info}{Professor Yiannis Demiris} and \href{https://maththrills.com}{Professor Michael Milford}. Furthermore, I would like to thank the sponsors of this research; research was funded in part by by the EU FP7 project WYSIWYD under Grant 612139, the Samsung Global
Research Outreach program, the EU Horizon 2020 Project PAL (643783-RIA), the Australian Government via grant AUSMURIB000001 associated with ONR MURI grant N00014-19-1-2571, Intel Research via grant RV3.248.Fischer, and the Queensland University of Technology (QUT) through the Centre for Robotics.

\bibliographystyle{myplainnat}
\bibliography{references}
\end{document}